\documentclass{article} 
\usepackage{iclr2021_conference,times}
\usepackage{lineno}

\usepackage{comment}
\usepackage{hyperref}
\usepackage{graphicx}
\usepackage{url}
\usepackage{multirow}
\usepackage{placeins}
\usepackage{amsfonts}
\usepackage{subfig}
\usepackage{slashbox}

\usepackage{amsmath,amsfonts,bm}
\title{\center{Improving generalization by mimicking the human visual diet}}

\author{Spandan Madan$^\dagger$\\
  Harvard University\\
   \And
  You Li\\
  UW Madison\\
   \\
  \And 
  Mengmi Zhang \\
 NTU and A$^*$\text{STAR}, Singapore\\
  \And
  Hanspeter Pfister\\
  Harvard University\\
  \And
  Gabriel Kreiman$^\dagger$\\
  Harvard University\\
}

\iclrfinalcopy 
\begin{document}
\setlength{\textwidth}{14.25cm}

\maketitle

\newcommand\blfootnote[1]{%
  \begingroup
  \renewcommand\thefootnote{}\footnote{#1}%
  \addtocounter{footnote}{-1}%
  \endgroup
}

\newcommand\extrafootertext[1]{%
    \bgroup
    \renewcommand\thefootnote{\fnsymbol{footnote}}%
    \renewcommand\thempfootnote{\fnsymbol{mpfootnote}}%
    \footnotetext[0]{#1}%
    \egroup
}

\extrafootertext{$^\dagger$ \emph{Corresponding authors:} \textit{spandan\_madan@seas.harvard.edu, gabriel.kreiman@tch.harvard.edu}}

\newcommand*{\eg}{\emph{e.g.,~}}
\newcommand*{\ie}{\emph{ie.,~}}
\newcommand*{\etal}{\emph{et al.~}}
\newcommand*{\etc}{\emph{etc.}}

\begin{abstract}
We present a new perspective on bridging the generalization gap between biological and computer vision---mimicking the human visual diet. While computer vision models rely on internet-scraped datasets, humans learn from limited 3D scenes under diverse real-world transformations with objects in natural context. Our results demonstrate that incorporating variations and contextual cues ubiquitous in the human visual training data (visual diet) significantly improves generalization to real-world transformations such as lighting, viewpoint, and material changes. This improvement also extends to generalizing from synthetic to real-world data---all models trained with a human-like visual diet outperform specialized architectures by large margins when tested on natural image data. These experiments are enabled by our two key contributions: a novel dataset capturing scene context and diverse real-world transformations to mimic the human visual diet, and a transformer model tailored to leverage these aspects of the human visual diet. All data and source code can be accessed at \url{https://github.com/Spandan-Madan/human_visual_diet}.

\end{abstract}


\section{Main}\label{sec:introduction}
Biological vision generalizes effortlessly across real-world transformations including changes in lighting, texture, and viewpoint. 
In contrast, visual recognition models are well known to fail at generalizing across real-world transformations including 2D rotations and shifts~\cite{zhang2019making, chaman2021truly}, changes in lighting~\cite{madan2021small,beery2018recognition, zhang2021adversarial}, object viewpoints~\cite{barbu2019objectnet, liu2018beyond, zeng2019adversarial, madan2021small, sakai2022three}, and color changes~\cite{joshi2019semantic,shamsabadi2020colorfool}, among others.  The dominant narrative guiding recent approaches to bridge the generalization gap is to improve \textit{how} machines process the data provided to them. Using large-scale internet-scraped datasets as benchmarks, modern approaches have attempted to improve pre-processing through data augmentation~\cite{yun2019cutmix, hendrycks2019augmix, zhang2017mixup}, generative modeling~\cite{ilse2020diva, wang2020cross}, extracting features better suited for generalization~\cite{muandet2013domain, li2018domain, li2018deep, motiian2017unified, shao2019multi, wang2021respecting}, proposing specialized architectures optimized for domain generalization~\cite{shahtalebi2021sand, DBLP:journals/corr/SunS16acoral, arjovsky2019invariant, kim2021selfreg, vedantam2021empirical, krueger2021out, blanchard2017domain}, or using specialized models to detect out-of-distribution (OOD) samples to process them separately~\cite{ren2019likelihood, shama2019detecting, hodge2004survey, aggarwal2001outlier}, among others. However, despite unprecedented progress in these closely related fields of domain adaptation, domain generalization, and out-of-distribution detection, human-like generalization remains an unsolved problem.

In this article, we present an alternative perspective on addressing this generalization gap inspired by how humans and other animals learn---instead of focusing on differences in \textit{how }data are processed, we focus on the fundamental differences in the data, i.e., the \emph{visual diet} of humans and machines. It is well documented that data fed during training can have profound impacts on the behaviour of both biological~\cite{kandel2000principles, kreiman2021biological, arcaro2017seeing, hubel1964effects, daw1976kittens, wood2018development, wood2022development} and computer vision~\cite{madan2022and, sakai2022three, madan2021small, orhan2020self}. Here, we investigate how generalization behaviour of visual recognition models changes as they are trained with a human-like visual diet, as opposed to internet-scraped datasets which are currently at the heart of most modern computer vision models.

\begin{figure}[t!]
    \centering
    \includegraphics[width=0.95\textwidth]{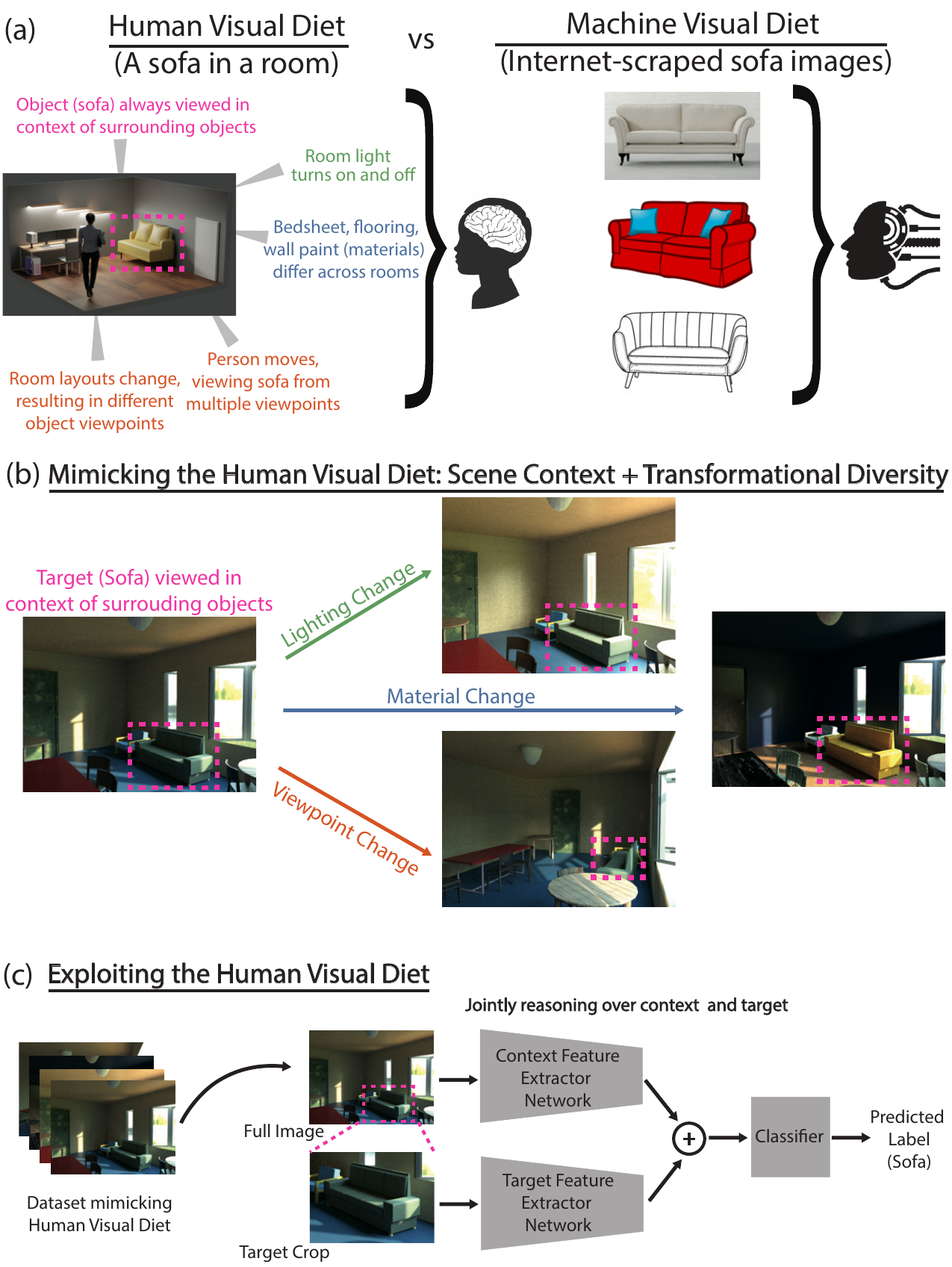}
    \caption{\textbf{Mimicking and exploiting the human visual diet.} (a) Comparing human and machine visual diets: The desk in the 3D room is viewed under a variety of real-world transformations which are essential components of the human visual diet. Furthermore, objects are always seen in context of their surroundings. In contrast, sample images of internet-scraped desks which constitute the machine visual diet do not contain these real-world transformations, or scene context. (b) Mimicking the human visual diet by introducing disentangled lighting, material, and viewpoint changes to a 3D scene where objects are shown in context. (c) Exploiting the human visual diet by using a two-stream architecture which reasons over both target object and its surrounding scene context.
    }
    \label{fig:fig_1_overview}
\end{figure}

\begin{figure}[t!]
    \centering
    \includegraphics[width=0.8\textwidth]{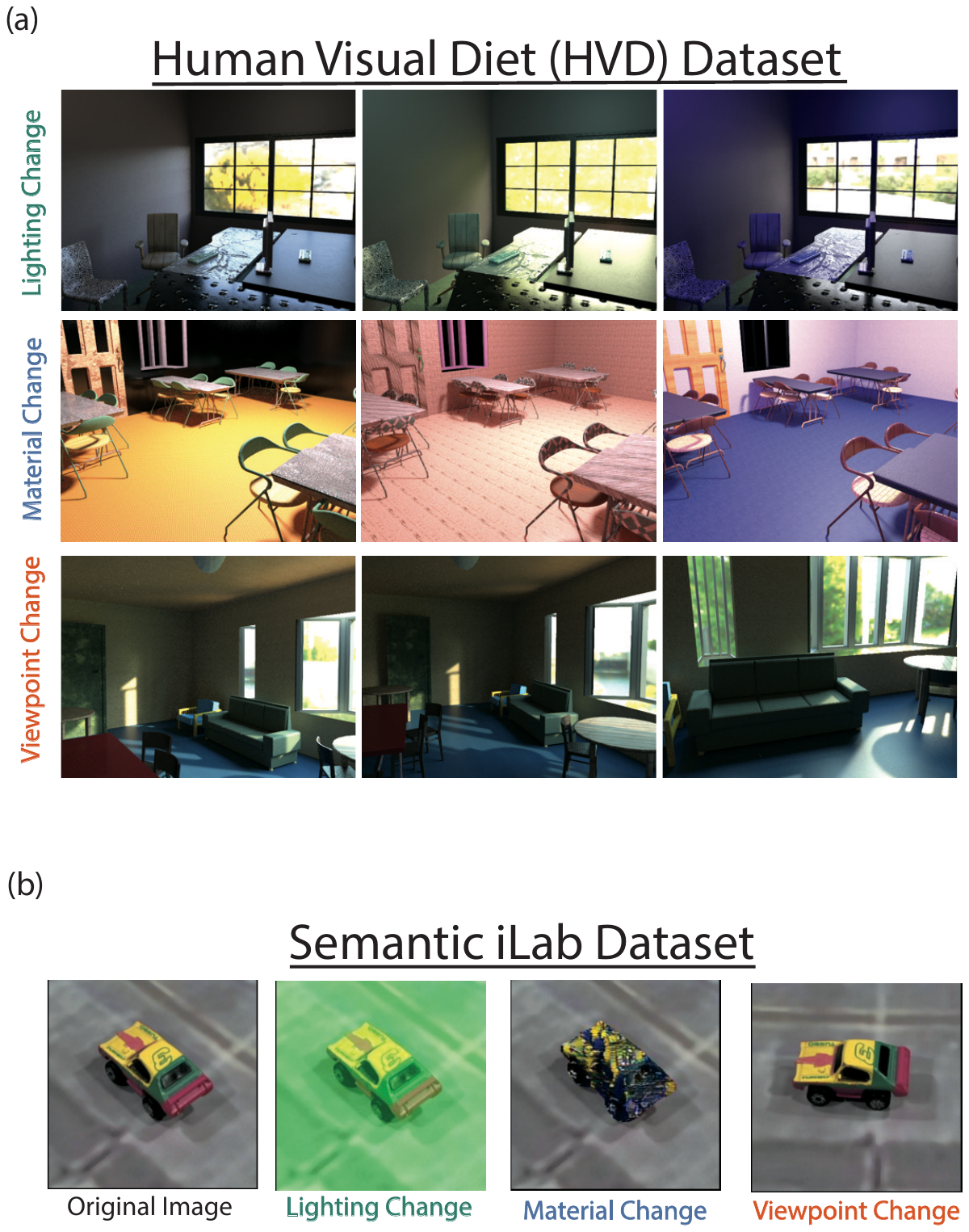}
    \caption{\textbf{Datasets with real-world transformations.} (a) Sample images from the Human Visual Diet dataset: We created
15 photo-realistic domains with three, disentangled real-world transformations---lighting, material, and viewpoint changes. Each 3D scene was created by reconstructing an existing ScanNet~\cite{dai2017scannet} scene using the OpenRooms framework~\cite{li2020openrooms}, followed by introduction of controlled changes in scene parameters before rendering these images. (b) Sample images from the Semantic-iLab dataset: We modify the existing iLab dataset~\cite{borji2016ilab} by augmenting images with changes in lighting and material. These changes are achieved by modifying the white balance and using AdaIN~\cite{huang2017adain} based style transfer, respectively.} 
    \label{fig:fig_2_datasets}
\end{figure}

\textbf{Figure ~\ref{fig:fig_1_overview}(a)} highlights two major differences between the typical human and machine visual diets. First, children learn from the physical space they occupy---a few $3$D scenes and objects while sampling densely from their surroundings under diverse real-world transformations. This includes viewing the same room from different viewpoints, under different lighting over the day, natural occlusions due to the room's layout, and changes in object textures. Second, humans rarely see objects in complete isolation but rather encounter objects in the context of their surroundings. In contrast, popular internet scraped datasets like ImageNet~\cite{deng2009imagenet} and CIFAR~\cite{krizhevsky2009learning} contain sparsely sampled information ($\sim 1$ photograph) from a large number of scenes. Each image is a single snapshot into the information available in the original scene, and data augmentation methods like $2$D rotations and crops do not reflect the complex real-world transformations seen by the human visual system. Furthermore, these images often contain objects with minimal scene context and most of the image is occupied by single objects placed in the center. For brevity, we refer to these as differences in \textit{transformational diversity} and \textit{scene context}, respectively.




 Our main finding is that mimicking and leveraging a human-like visual diet enables better generalization, and models trained with such a diet outperform specialized architectures trained without transformational diversity and scene context. The experiments in this work are enabled by two key technical contributions. First, we introduce a new \textbf{ Human Visual Diet (HVD)} dataset, which models both transformational diversity and scene context to better mimick data seen by the human visual system. \textbf{Figure ~\ref{fig:fig_2_datasets}} showcases how HVD was created. Second, we propose a new model to leverage the visual diet presented in HVD, as existing recognition models are not designed to exploit a human-like diet. We call this model the \textbf{Human Diet Network (HDNet)}. As shown in \textbf{Fig.~\ref{fig:fig_1_overview}(c)}, HDNet uses a two-stream architecture where one stream operates on the target object, while the other stream operates on the scene context to jointly reason over target and context to perform visual recognition. HDNet also utilizes transformational diversity by employing a contrastive loss over real-world transformations in the form of lighting, material, and viewpoint changes (\textbf{Section ~\ref{sec:methods}}). With these findings, we present compelling evidence for the field of computer vision to move towards biologically inspired, diverse data which more accurately model the training data seen by human vision.

\section{Results}\label{sec:results}
We evaluated the utility of mimicking the human visual diet in improving generalization across real-world transformations in the form of lighting, material and viewpoint changes. First, we evaluated the generalization capabilities of standard architectures trained with data simulating the visual diet commonly used in computer vision---consisting of large-scale, internet-scraped data with low real-world transformational diversity and
minimal scene context (\textbf{Sec.~\ref{subsec:viewpoint_material_light}}). Then, we confirmed that incorporating and utilizing these two hallmarks of the human visual diet (real-world transformational diversity and scene context) significantly improves generalization (\textbf{Sec.~\ref{subsec:visual_diet_to_rescue}}). We also confirmed that this approach outperforms specialized domain generalization architectures trained with alternatives like traditional data augmentation methods, and specialized GAN-based augmentation methods (\textbf{Sec.~\ref{subsec:augmentation_vs_diversity}}). Finally, as a real litmus test for our findings, we demonstrated that utilizing a human-like visual diet improves generalization from synthetic training data to real-world, natural image data (\textbf{Sec.~\ref{subsec:generalizing_to_real_world}}).


These experiments are enabled by two datasets containing real-world transformations---the human visual diet (HVD) dataset which is introduced here, and the Semantic-iLab dataset which we created by modifying the multi-view iLab dataset~\cite{borji2016ilab}. Each dataset was created to consist of $15$ domains with disjoint real-world transformations in the form of lighting, material and viewpoint changes ($5$ domains per transformation). Sample images from these datasets are shown in \textbf{Fig.~\ref{fig:fig_2_datasets}}, and Supplementary \textbf{Fig.~\ref{fig:light_shift}}, \textbf{Fig.~\ref{fig:material_shift}} and \textbf{Fig.~\ref{fig:viewpoint_shift}}. For instance, the $5$ material domains in HVD were created by starting with $250$ object materials and splitting them into $5$ non-overlapping sets of $50$ each. For each material domain, these $50$ materials were randomly assigned to scene objects, and images were rendered. This procedure ensures that every material shows up in only one particular domain while still ensuring high material diversity across scenes. For every transformation, one domain was held out for testing (\eg out-of-distribution (OOD) Materials), and never used for training any model. Data diversity is defined as the number of domains the training data is sampled from (ranging from $1$ to $4$). A similar protocol was used to create disjoint domains and study generalization across lighting and viewpoint changes in the HVD dataset. Experiments with the Semantic-iLab dataset also followed the same protocol as well---$15$ domains in all with $5$ domains per transformation, and models trained with $1,2,3$ or $4$ domains with $1$ held-out domain used for testing which was never used for training (\textbf{Sec.~\ref{sec:methods}}).


\begin{figure}[t!]
    \centering
    \includegraphics[width=0.75\textwidth]{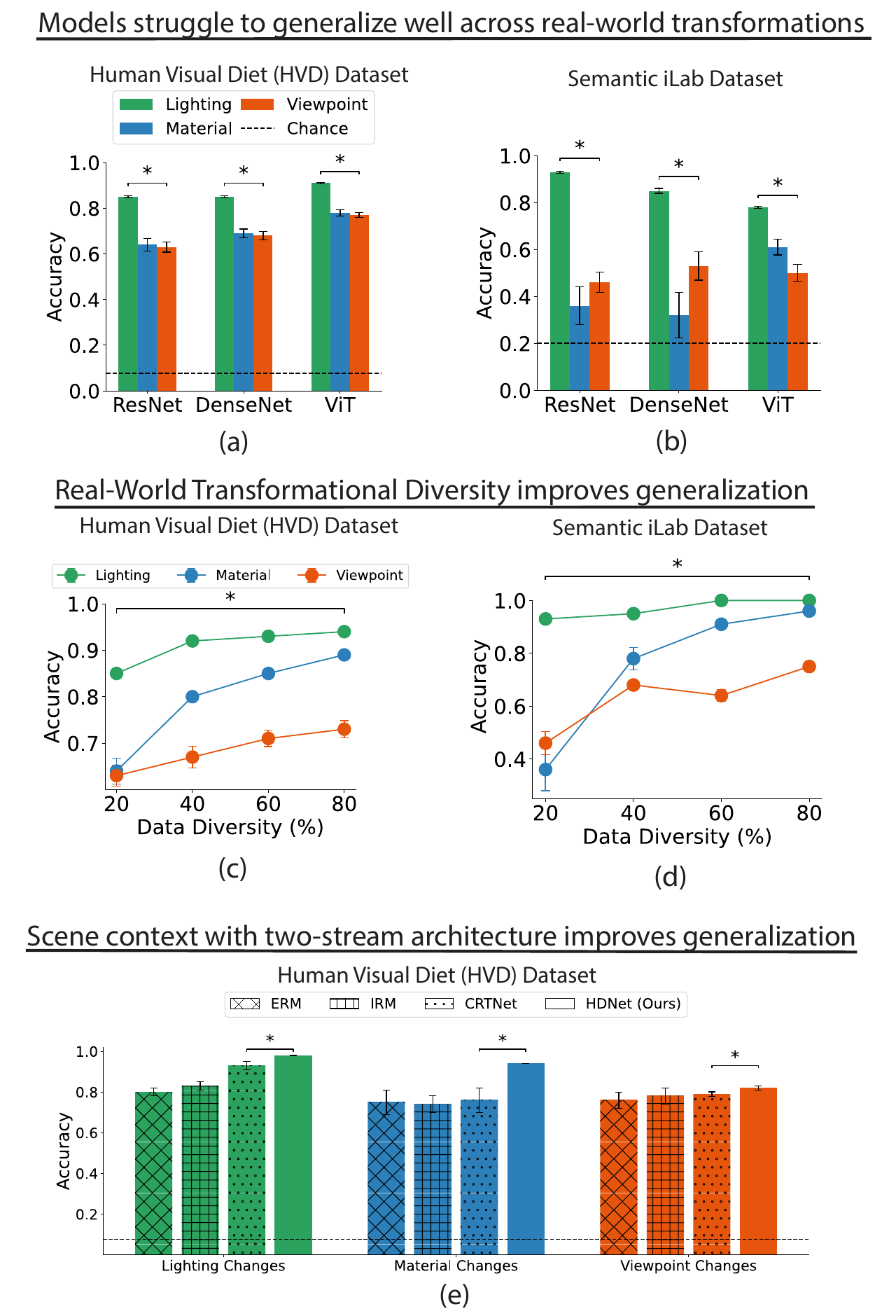}
    \caption{\textbf{Human Visual Diet leads to significantly improved generalization across real-world transformations.}((a) Existing models struggle to generalize across real-world transformations, especially material and viewpoint changes. This result holds for both HVD and Semantic-iLab datasets. (b) Increasing real-world transformational diversity leads to a significant increase in generalization performance for all transformations (lighting, viewpoint and materials) for both datasets. (c) HDNet leverages scene context resulting in substantially better generalization than seminal domain generalization architectures like ERM~\cite{blanchard2017domainmtl}, IRM~\cite{arjovsky2019invariant}. HDNet is designed to incorporate scene context into visual recognition, by using a two-stream architecture to reason over the target object and scene context simultaneously. In contrast, above mentioned state-of-the-art approaches for domain generalization are single stream architectures that do not leverage scene context. HDNet also beats a suite of additional domain generalization baselines presented in Table~\ref{table:dg_benchmarks}. The closest performing baseline is another context-aware ~\cite{kim2021selfreg} model (CRTNet\cite{bomatter2021pigs}), and our proposed model beats theses baselines for all three transformations with statistical significance. For all plots, statistical significance is evaluated using a two-sample t-test, and an $^*$ indicates a p-value lower than the threshold of $0.05$. See methods for additional details.} 
    \label{fig:fig_3_main_results}
\end{figure}


\subsection{Models trained without human-like visual diet struggle to generalize across real-world transformations}\label{subsec:viewpoint_material_light}
We simulated the typical visual diet used in many computer vision studies consisting of large-scale, internet-scraped datasets by training visual recognition models with low real-world transformational diversity and minimal scene context. Specifically, for each type of transformation (lighting, materials, and viewpoints), models were trained with only $1$ domain and then tested on the held-out test domain with the corresponding transformation. Minimal scene context was simulated by training on cropped images showing only the target object as is common in visual recognition datasets like ImageNet~\cite{deng2009imagenet}. All models presented below were pre-trained on ImageNet.

We evaluated how well a ResNet18 \cite{he2016deep} trained with one lighting domain and minimal context could generalize to the held-out, unseen lighting domain. We similarly evaluated how well a ResNet could generalize across different material domains, or viewpoint domains. 
For HVD, \textbf{Fig.~\ref{fig:fig_3_main_results}(a)} shows that ResNet generalized better across lighting changes than material changes (two-sided t-test, $p<10^{-5}$) or viewpoint changes (two-sided t-test, $p<10^{-6}$). These results show that there is ample room for improvement for models trained with a visual diet typical in computer vision, especially when tested on unseen material and viewpoint transformations. Similar conclusions can also be drawn  for DenseNet~\cite{huang2017densely} and ViT~\cite{dosovitskiy2020image} architectures.

We also confirmed that the above findings extend to the natural image Semantic-iLab dataset (\textbf{Fig.~\ref{fig:fig_3_main_results}(b)}). ResNet generalized better across lighting changes than material changes (two-sided t-test, $p<10^{-6}$) or viewpoint changes (two-sided t-test, $p<10^{-6}$). In the Semantic-iLab dataset, the degree of generalization for material and viewpoints were particularly low. The same conclusions held true for DenseNet and ViT architectures as well.

The generalization deficits were even more striking for the Semantic-iLab dataset compared to the HVD dataset. One potential explanation for this is that unlike the highly photorealistic HVD dataset which was rendered with complete control over lighting and material, Semantic-iLab was created by introducing approximate light and material changes to images in the iLab~\cite{borji2016ilab} dataset using white balance modifications and style transfer which result in stark changes across domains (see \textbf{Sec.~\ref{sec:methods}} for details). In sum, state-of-the-art computer vision architectures trained with minimal transformational diversity show only moderate generalization across real-world object transformations, especially for material and viewpoint changes.

\subsection{A human-like visual diet improves generalization}\label{subsec:visual_diet_to_rescue}
Below we present evidence in support of our main hypothesis---mimicking and utilizing a human-like visual diet improves this generalization behavior. Specifically, we study two aspects of the human visual diet---Real-World Transformational Diversity (RWTD) and Scene Context, as presented in Fig.~\ref{fig:fig_1_overview}. The impact of these two factors are reported in \textbf{Sec.~\ref{subsubsec:transformational_diversity}} and \textbf{Sec.~\ref{subsubsec:scene_context}} respectively.

\subsubsection{Real-word transformational diversity improves generalization}\label{subsubsec:transformational_diversity}

We evaluated the effect of increasing transformational diversity on generalization by ResNet. 
Real-World Transformational Diversity (RWTD) shown to the model during training is defined by the number of domains from which the training dataset was sampled. For each transformation (lighting, material, or viewpoint), there was one fixed testing domain which stayed constant across all models. The training dataset was always of a fixed size, but the training images were sampled from $1, 2, 3$ or $4$ domains (corresponding to 20\%, 40\%, 60\%, and 80\% data diversity). This procedure resulted in different levels of transformational diversity shown during training while maintaining a fixed dataset size, thus disentangling the role of transformational diversity and dataset size (\textbf{Sec.~\ref{sec:methods}}). 

Performance monotonically increased with data diversity in the HVD dataset for all three transformations (\textbf{Fig.~\ref{fig:fig_3_main_results}(c)}: lighting: $0.85$ to $0.94$, $p<10^{-6}$; material: $0.64$ to $0.89$, $p<10^{-5}$; viewpoint: $0.63$ to $0.73$, $p<10^{-6}$). The improvement in generalization accuracy with data diversity was significantly greater for unseen materials than for unseen lighting ($p<10^{-4}$) and unseen viewpoints ($p<10^{-4}$). The smaller increase for unseen lighting changes is expected as the performance was already high for this transformation with low data diversity which left less room for improvement. For unseen viewpoints, these experiments suggest that despite a statistically significant improvement, increasing data diversity is not a sufficient solution for solving generalization, consistent with past work investigating invariance to 3D viewpoint changes~\cite{madan2021small, sakai2021three, cooper2021out, siddiqui2023investigating}.

%
To ensure our findings also extend to natural image data, we also report results with the Semantic-iLab dataset in \textbf{Fig.~\ref{fig:fig_3_main_results}(d)}. Increasing real-world transformational diversity improved generalization also for the Semantic-iLab dataset: lighting: $0.93$ to $1.0$,  $p<10^{-3}$; materials: $0.36$ to $0.96$, $p<10^{-4}$; viewpoint: $0.46$ to $0.75$, $p<10^{-7}$. As with the HVD dataset, improvement in generalization was higher for unseen materials than for unseen lighting ($p<10^{-3}$) and unseen viewpoints ($p<10^{-6}$).

The results across both datasets consistently show that generalization across real-world transformations increases with transformational diversity. With sufficient diversity, generalization to unseen lighting and materials reached almost ceiling levels. However, despite improvement, unseen viewpoints remain an open challenge.

\subsubsection{Leveraging scene Context substantially improves generalization to OOD real-world transformations.}\label{subsubsec:scene_context}
Next, we evaluated the impact of incorporating another hallmark of the human visual diet---scene context. As shown in the schematic in \textbf{Fig.~\ref{fig:fig_1_overview}(c)}, our proposed architecture, HDNet, uses a two-stream architecture that reasons over both the target object and the scene context to classify images. One stream operates on a crop around the target object, while the second stream operates on the full image containing the scene context. HDNet also performs an additional contrastive loss across real-world transformations in the form of lighting, materials, and viewpoints. While contrastive loss has become a staple in modern vision models, this is the first implementation applying contrastive loss over real-world transformations including lighting, materials and 3D viewpoint changes. These features allow HDNet to exploit both aspects of the human visual diet modeled in this study---transformational diversity and scene context (for additional details on the architecture, see \textbf{Sec.~\ref{sec:methods})}.


We compared results for multiple state-of-the-art algorithms including a suite of domain generalization methods, a recent context-aware model (CRTNet~\cite{bomatter2021pigs}), and a FasterRCNN model modified to perform visual recognition with our proposed HDNet which utilizes the human-like visual diet. For each real-world transformation, all models were trained with $80\%$ Transformational Diversity, i.e., $4$ training domains, and tested on the corresponding held-out test domain. Note that these domain generalization baselines are highly-specialized methodologies with novel engineering combining architectural modifications, optimization strategies, and model selection criteria optimized for the task of domain generalization. In contrast, HDNet is a general purpose architecture meant for visual recognition designed to leverage a human-like visual diet. 

Our hypothesis is that the benefits from leveraging the correct diet can outperform the gains from these specialized domain generalization methodologies. Results investigating this hypothesis are presented in \textbf{Fig.~\ref{fig:fig_3_main_results}}(e) and Table~\ref{table:dg_benchmarks}. As shown in \textbf{Fig.~\ref{fig:fig_3_main_results}}(e), HDNet beat ERM~\cite{gulrajani2020search} across all three real-world transformations with an accuracies of $0.98, 0.94, 0.83$ compared to ERM's $0.83, 0.75, 0.79$ on unseen lighting, material, and viewpoint changes respectively. Similarly, HDNet also beat IRM~\cite{arjovsky2019invariant}) which achieves lower Top-1 accuracies of $0.83, 0.74, 0.79$ on unseen lighting, material, and viewpoint changes respectively. The best performing baseline was found to be another context-aware model---CRTNET~\cite{bomatter2021pigs}) which is also reported in \textbf{Fig.~\ref{fig:fig_3_main_results}}(e). Our proposed HDNet model beat this CRTNet baseline with statistical significance on all three real-world transformations. For unseen lighting, HDNet beats CRTNet with an accuracy of $0.98$ compared to $0.93$ (two-sided t-test, $p<0.05$). For unseen material changes, HDNet achieves $0.94$ which is higher than HDNet's Top-1 accuracy of $0.76$ (two-sided t-test, $p<0.05$). Similarly, for unseen viewpoint changes, HDNet beats MTL with an accuracy of $0.83$ compared to $0.79$ (two-sided t-test, $p<0.05$). In summary, both context-aware models (our proposed HDNet and existing CRTNet) outperformed specialized domain generalization approaches on all real-world transformations, and our proposed HDNet also outperformed the closest baseline (CRTNet) with statistical significance.

We also compared HDNet against a suite of additional baselines as reported in \textbf{Table~\ref{table:dg_benchmarks}}. This includes additional domain generalization benchmarks, and a modified FasterRCNN~\cite{ren2015faster} designed to do visual recognition~\cite{bomatter2021pigs}. Findings are consistent across all architectures---HDNet, which utilizes scene context, outperformed all benchmarks across all real-world transformations including lighting, material, and viewpoint changes.


\begin{table*}[t!]
\centering
\begin{tabular}{c|c|c|c|c|c|c|c||c}
    \hline
    \hline
    \multirow{3}{*}{\parbox{2.5cm}{\centering \textbf{Real-World Transformation}}} 
     & \multirow{3}{*}{\parbox{0.8cm}{\centering AND Mask \cite{shahtalebi2021sand}}}& \multirow{3}{*}{\parbox{0.8cm}{\centering CAD \cite{blanchard2017domain}}} & \multirow{3}{*}{\parbox{0.8cm}{\centering COR AL \cite{DBLP:journals/corr/SunS16acoral}}} & \multirow{3}{*}{\parbox{0.8cm}{\centering MTL \cite{blanchard2017domainmtl}}}  &  \multirow{3}{*}{\parbox{0.8cm}{\centering Self Reg \cite{kim2021selfreg}}} & \multirow{3}{*}{\parbox{0.8cm}{\centering VREx \cite{krueger2021out}}} & 
     \multirow{3}{*}{\parbox{1cm}{\centering Faster RCNN \cite{ren2015faster}}} & 
     \multirow{3}{*}{\parbox{1cm}{\centering HDNet (ours)}} \\
     & & & & & & & & \\
     & & & & & & & & \\
     \hline
     Light & 0.82 & 0.80 & 0.81 & 0.81 & 0.75 & 0.83 & 0.95 & \textbf{0.98}\\
     \hline
      Materials & 0.75 & 0.75 & 0.75  & 0.74 & 0.74 & 0.75 & 0.78 & \textbf{0.94}\\
      \hline
      Viewpoints & 0.75 & 0.77 & 0.79 & 0.79 & 0.76 & 0.78 & 0.65 & \textbf{0.83}\\
     \hline
\end{tabular}
\caption{\textbf{Contextual information improves generalization across OOD transformations.} We present additional baselines comparing HDNet to several domain generalization methods that do not use contextual cues, building on results presented in \textbf{Fig.~\ref{fig:fig_3_main_results}}. All architectures were trained with transformational diversity of 80\% and tested on the held-out remaining 20\% of the corresponding semantic shift. HDNet beats all baselines by a large margin for all semantic shifts. HDNet also beat a version of FasterRCNN modified to do object recognition, which was provided with contextual cues. The best performing model (HDNet) has been shown in boldface for all real-world transformations.
}
\label{table:dg_benchmarks}
\end{table*}



\begin{table}[h]
\centering
\begin{tabular}{c|c|c|c}
    \hline
    \hline
    \multirow{2}{*}{\parbox{2.5cm}{\centering \textbf{Real-World Transformation}}} & \multirow{2}{*}{\textbf{Architecture}} & \multirow{2}{*}{\textbf{1 Stream}} & \multirow{2}{*}{\textbf{2 Stream}}\\
    & & & \\
     \hline
     \multirow{4}{*}{Lighting} & ResNet & $0.85 \pm 0.004$ & ${0.95 \pm 0.009^*}$\\
      & ViT &  $0.91 \pm 0.003$ & ${0.97 \pm 0.007^*}$\\  
      & HDNet (Ours) &  - & $\mathbf{0.98 \pm 0.001}$\\  
      \hline
      \multirow{4}{*}{Materials} & ResNet &  $0.64 \pm 0.03$ & ${0.83 \pm 0.008^*}$\\
      &  ViT &  $0.78 \pm 0.01$ & ${0.92 \pm 0.003^*}$\\
      & HDNet (Ours) & - & $\mathbf{0.94 \pm 0.002}$\\  
      \hline
     \multirow{4}{*}{Viewpoint} & ResNet &  $0.63 \pm 0.02$ & ${0.72 \pm 0.009^*}$\\
      & ViT & $0.77 \pm 0.01$ & ${0.83 \pm 0.001^*}$\\
      & HDNet (Ours) &  - & $\mathbf{0.83 \pm 0.006 }$\\  
      \hline
\end{tabular}
\caption{\textbf{Adding scene context improves performance independent of architecture.} 
Following the design of HDNet shown in \textbf{Fig.~\ref{fig:fig_1_overview}(c)}, we modified standard architectures to have two streams---one operating on the target, and the other one on the contextual information. Representations for both streams are then concatenated and passed through a classification layer as shown in \textbf{Fig.~\ref{fig:fig_1_overview}(c)}. We train the standard one-stream and these modified two-stream architectures on HVD, and report the average Top-1 accuracy for all models . We also report error bars, which measures the variance in accuracies over categories. Both the ResNet and the ViT architectures lead to a large improvement in generalization for all semantic shifts when modified to leverage scene context. To ensure we study impact of context independent of data diversity, all models were trained on $4$ domains, i.e., $80\%$ transformational diversity and tested on the held out domain. Best performing model (HDNet) has been shown in boldface for all real-world transformations. A $*$ refers to statistically significant improvement in performance when using a two-stream architecture as compared to a one-stream architecture (two-sided t-test, $p<0.05$).
}\label{table:control_1}
\end{table}
Given the success of HDNet, we asked whether implementing a two-stream separation of target and context would also improve performance for other architectures. 
We modified ResNet18~\cite{he2016deep} and ViT~\cite{dosovitskiy2020image} to leverage scene context in the same way as HDNet. For ResNet, a two-stream version was made where each stream is a ResNet backbone. One stream operates on the target, and the other one on the scene context. Output features from each stream were concatenated, and passed through a fully connected layer for classification as shown in \textbf{Fig.~\ref{fig:fig_1_overview}(c)}. The two-stream architecture for ViT was analogous. In contrast, the one-stream architecture did not use scene context and operated on the target object alone (see \textbf{Sec.~\ref{sec:methods}} for additional details). 

The two-stream architectures consistently led to improved performance with statistical significance (two-sided t test, $p<0.05$), as shown in \textbf{Table~\ref{table:control_1}}.
The increase in performance is due to the addition of contextual information and not to the two-stream architecture \emph{per se}. Indeed, when both streams were trained with the target information, there was a decrease in performance (\textbf{Table.~\ref{table:control_3}}). 


\begin{table}[h]
\centering
\begin{tabular}{c|c|c|c}
\hline
\hline
\multirow{3}{*}{\parbox{1.5cm}{\centering Semantic Shift}} & \multirow{3}{1.2cm}{\centering Full Context ($\sigma=0$)} & \multirow{3}{1.7cm}{\centering Less Context ($\sigma=25$)} & \multirow{3}{1.7cm}{\centering Least Context ($\sigma=125$)}\\
& & &\\
& & &\\
\hline

 Lighting & $\mathbf{0.98 \pm 0.001} $ & $0.96 \pm 0.001$ & $0.94 \pm 0.001$\\
 \hline
 Material & $\mathbf{0.94 \pm 0.002}$ & $0.88 \pm 0.01$ & $0.83 \pm 0.006$\\
 \hline
 Viewpoint & $\mathbf{0.83 \pm 0.006}$ & $0.77 \pm 0.01$ & $0.76 \pm 0.01$\\
 \hline
\end{tabular}
\caption{\textbf{Blurring scene context worsens generalization performance}. We trained and tested HDNet with the scene context in HVD images blurred using a Gaussian blur. Here, $\sigma$ is the standard deviation for the gaussian kernel applied to the image as a filter. Thus, blurring increases with $\sigma$. We applied three values for $\sigma$---$0$,$25$, and $125$. For brevity, numbers less than 0.001 are reported as 0.001.}

\label{table:control_2}
\end{table}

To further understand the role of contextual information on visual recognition, we evaluated the impact of blurring the scene context while keeping the target intact \cite{zhang2020putting}. For each real-world transformation, we trained and tested models with increasing levels of Gaussian blurring applied to the scene context. Blurring was applied to the images in the form of a Gaussian kernel filter, with the kernel standard deviation ($\sigma$) set to $0$, $25$, or $125$. The cropped image of the target object was passed to the second stream of the network without blurring (Supplementary \textbf{Sec.~\ref{supplesec:additional_controls}}). There was a drop in performance with context blurring for all three real-world transformations(\textbf{Table~\ref{table:control_2}}).

\subsection{Alternatives to mimicking the human-like visual diet are not equally advantageous}~\label{subsec:augmentation_vs_diversity}

\begin{figure}[t!]
    \centering
    \includegraphics[width=0.65\textwidth]{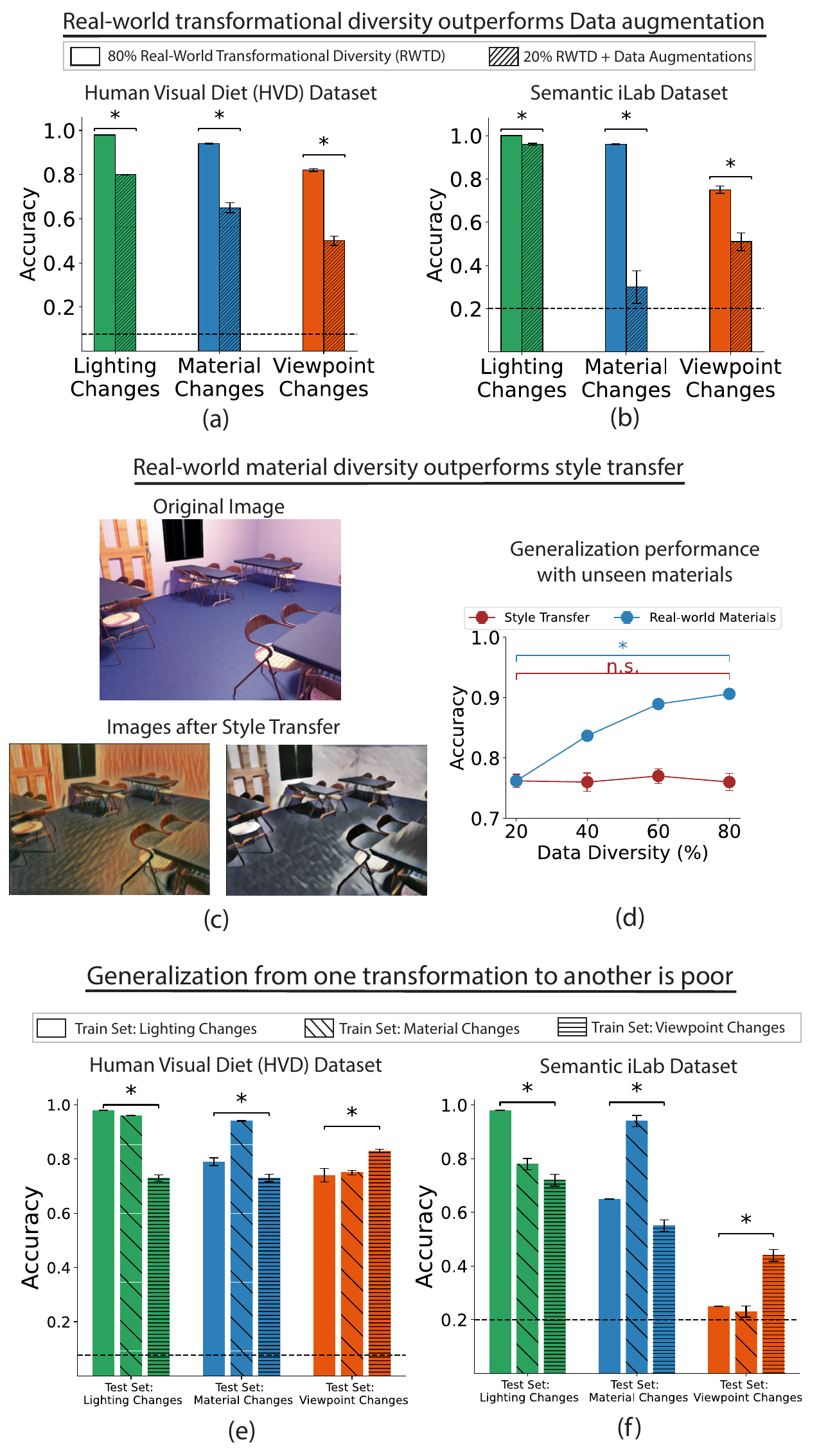}
    \caption{\textbf{Data post-processing does not match gains from collecting data mimicking the human visual diet}. (a) Models trained 80\% real-world transformational diversity (RWTD) significantly outperform modesl trained with 20\% along with
    traditional data augmentation. This is true for all transformations (lighting, material, and viewpoint) across both HVD and Semantic-iLab datasets. Number of images is held constant in these experiments. (b) Sample images from style transfer domains created using AdaIn~\cite{huang2017adain}, alongside accuracies of models trained with these domains. Models trained on style transfer domains generalize significantly worse than those trained with material diversity. (c) Generalization from one transformation to another (asymmetric diversity) does not help as much as training with the correct transformation---best generalization to unseen materials is achieved when material diversity is added to the training data. For generalizing to unseen light and viewpoint changes as well, training with the corresponding real-world diversity helps the most.}
    \label{fig:fig_4_alternate_methods}
\end{figure}

We asked whether similar performance gains to the ones achieved by using a human-like visual diet could be obtained by alternate strategies that can be implemented on existing internet-scraped datasets. Achieving comparable performance using internet-scraped datasets would bypass the laborious efforts in collecting data and adequate controls. We investigated three such approaches which have been popularly used in the literature---data augmentation, using generative models to modify existing images, and increasing diversity on more easily controllable real-world transformations. 

\subsubsection{Traditional Data augmentation does not provide similar performance to true diversity with real-world transformations.}

Here we investigated how real-world transformational diversity (RWTD) compares to traditional data augmentation strategies including 2D rotations, scaling, and changes in contrast. Models trained with a visual diet consisting of $80\%$ RWTD were reported in \textbf{Fig.3(e)}. We compared these with models trained with a visual diet consisting of $20\%$ RWTD + traditional augmentation. As before, all models were tested on unseen lighting, material, and viewpoint changes. 

Note that the number of training images was kept constant across all training scenarios to evaluate the quality of the training images rather than their quantity. Training set size equalization was achieved by sampling fewer images per domain in the $80\%$ RTWD training set. For instance, for HVD experiments with unseen viewpoints we sampled $15,000$ training images per viewpoint domain to construct the training set with $20\%$ RWTD + Data Augmentations. In comparison, we sampled only $3,750$ per viewpoint domain to construct the $80\%$ RWTD training set. Thus, the initial sizes of the 80\%RWTD and the 20\%RWTD+Data Augmentation training sets was identical. However, due to data augmentations being stochastic the total number of unique images shown to models trained with data augmentations was much larger. Assuming a unique image was created by data augmentation in every epoch, over 50 epochs the dataset size would be 50 times larger with data augmentations. Additional details on dataset construction can be found in the methods in \textbf{Sec.~\ref{sec:methods}}.

HDNet trained on HVD with $80\%$ RWTD outperformed the same architecture trained with $20\%$ RWTD+traditional data augmentation for lighting changes (two-sided t test, $p<10^{-4}$), material changes (two-sided t test, $p<10^{-5}$), and viewpoint changes (two-sided t test, $p<10^{-6}$) (\textbf{Fig.~\ref{fig:fig_4_alternate_methods}(a)}). Similar conclusions were reached for the Semantic-iLab dataset. A ResNet model trained with $80\%$ RWTD outperformed the same architecture trained with $20\%$ RWTD+traditional data augmentation for lighting changes (two-sided t test, $p<10^{-4}$), material changes (two-sided t test, $p<10^{-7}$), and viewpoint changes (two-sided t test, $p<10^{-5}$) (\textbf{Fig.~\ref{fig:fig_4_alternate_methods}(b)}). 

One explanation for this finding could be that traditional data augmentation largely involves 2D affine operations (crops, rotations) or image-processing based methods (contrast, solarize) which are not necessarily representative of real-world transformations. In summary, the positive impact of a visual diet consisting of diverse lighting, material, and viewpoint changes (real-world transformational diversity) cannot be replicated by using traditional data augmentation applied to the dataset after data collection---diversity must be ensured at the data collection level.

\subsubsection{Real-world transformations outperform modifying a low diversity sample with generative models.}


Several existing works rely on increasing data diversity using AdaIn-based methods ~\cite{huang2017adain,zhou2021domain}. These style transfer methods change the colors in the image while retaining object boundaries, but do not modify materials explicitly as done in our HVD dataset. We evaluated how well models perform if diversity is increased using style transfer as opposed to material diversity. We started with one material domain, and created four additional domains using style transfer. Sample images of style transfer domains are shown in \textbf{Fig.~\ref{fig:fig_4_alternate_methods}(c)}. Corresponding images from the HVD dataset with real-world transformation in materials can be seen in \textbf{Fig.~\ref{fig:fig_2_datasets}(a)}. The total number of domains (and images) created using style transfer was kept the same as the material domains in HVD. The only difference in the training data was that instead of four additional material domains, we have four additional style transfer domains. We compare models trained with these two different visual diets---one consisting of four material domains, and the other consisting of four style transfer domains. All models are then tested on the same held out OOD Materials domain.

Style transfer domains did not enable models to generalize to new materials as well as the material shift domains presented in HVD
(\textbf{Fig.~\ref{fig:fig_4_alternate_methods}(d)}). 
These experiments support the notion that in order to build visual recognition models that can generalize to unseen materials, it is important to explicitly increase diversity using additional materials at the time of training data collection. The impact of diverse materials cannot be replicated by using style transfer to augment the dataset after data collection.



\subsubsection{Importance of diversity on all real-world transformations}
Some real-world transformations are easier to capture than others. For instance, capturing light changes during data collection might be significantly easier than collecting all possible room layouts, or object viewpoints. Thus, it would be beneficial if training with one transformation (\eg light changes) can improve performance on a different transformation (\eg viewpoint changes). We refer to such a regime as \textit{assymetric diversity}---as models are trained with one kind of diversity, and tested on a different kind of diversity (\textbf{Fig.~\ref{fig:fig_4_alternate_methods}(e),(f))}.

In all cases, the best generalization performance was obtained when training and testing with the same real-world transformation for both HVD (\textbf{Fig.~\ref{fig:fig_4_alternate_methods}(e)}) and Semantic-iLab datasets (\textbf{Fig.~\ref{fig:fig_4_alternate_methods}(f)}). In most cases, there was a drop in performance of $10\%$ or more when training in one transformation and testing in with a different (assymetric) transformation.
These experiments imply that to build models that generalize well, it is important to collect training data with multiple real-world transformations.


\subsection{Models trained with a human-like visual diet can generalize to real-world images}\label{subsec:generalizing_to_real_world}
\begin{figure}[!t]
    \centering
    \includegraphics[width=0.8\textwidth]{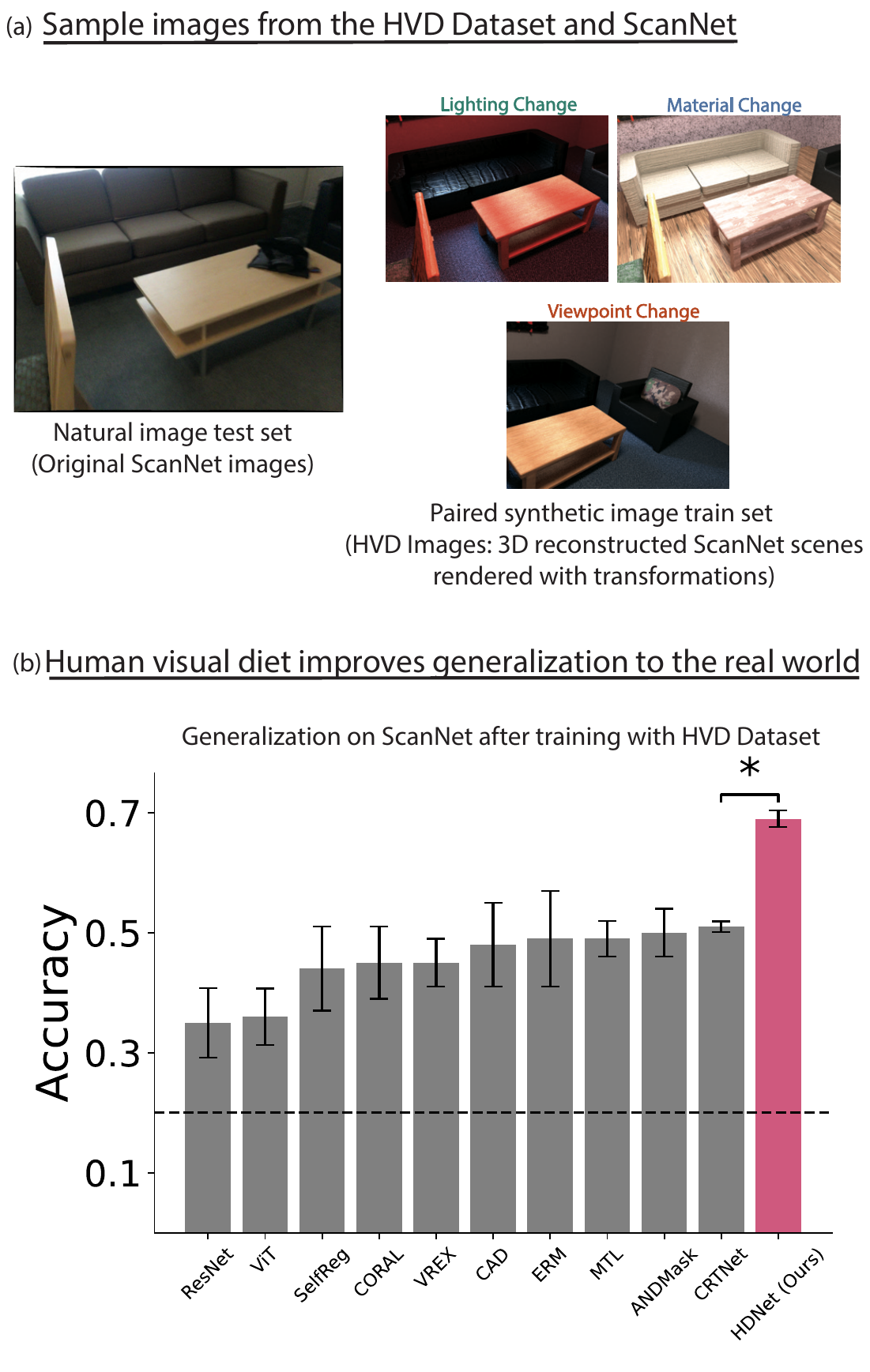}
    \caption{\textbf{Utility of the human visual diet in generalizing from synthetic to real-world, natural image data}. (a) Sample synthetic images from the HVD dataset used for training the model, and the corresponding real-world natural image from ScanNet used for testing. (b) Human Visual Diet enables substantially better generalization from synthetic to natural image data. Our approach, which mimics and effectively utilizes transformational diversity and scene context leads to better performance than all other baselines.
    }
    \label{fig:fig_5_natural_image_results}
\end{figure}
As a litmus test for our findings, we investigated if models trained with a human-like visual diet can generalize well to the natural images from the ScanNet dataset despite only being trained with synthetic images from HVD. \textbf{Fig.~\ref{fig:fig_5_natural_image_results}(a)} shows paired testing and training data across these two datasets. The test set is composed of natural images from the ScanNet dataset, while the training set consists of only synthetic images from HVD. HVD images were created by 3D reconstructing ScanNet scenes, and then rendering them under diverse lighting, material and viewpoint change as shown in\textbf{ Fig.~\ref{fig:fig_5_natural_image_results}(a)}.

We made three adaptations for these experiments. Firstly, as both ScanNet and ImageNet contain natural images and overlapping categories, we trained models from scratch to ensure pre-training does not interfere with our results. Thus, these models never saw any real-world images, not even ImageNet as they were not pretrained on those datasets. Secondly, we trained and tested models on overlapping classes between HVD and ShapeNet. Finally, we used the LabelMe~\cite{wada2018labelme} software to manually annotate a test set from ScanNet and training set for the HVD dataset using the same procedure to make sure biases from the annotation procedure do not impact experiments.


Thus, all models were trained purely on synthetic data from HVD and tested on only real-world natural image data from ScanNet as shown in \textbf{Fig.~\ref{fig:fig_5_natural_image_results}(a)}. While all the models that we tested performed above chance levels when tested in the real-world ScanNet images, there were large performance differences among models (\textbf{Fig.~\ref{fig:fig_5_natural_image_results}b}).

Results on generalization to the real world are presented in \textbf{Fig.~\ref{fig:fig_5_natural_image_results}}. As can be seen, the best performing model (IRM) trained without the human visual diet obtained classification performance of $0.51$, while HDNet trained with transformational diversity and scene context performs substantially better at $0.69$. Despite being trained only on synthetic data and no pre-training on natural images, HDNet generalizes well to the real world by leveraging attributes of the human visual diet. Our approach with the HVD dataset and HDNet, which mimics and utilizes two hallmarks of human visual diet (real-world transformational diversity and scene context) beats all benchmarks with statistical significance ($p<0.05$). This evidence confirms that our findings on Semantic-iLab and the HVD datasets also extend to real-world natural images from the ScanNet dataset.

\section{Discussion}\label{sec:discussion}
Alleviating the generalization gap between biological and computer vision remains an open and elusive problem. The past few years have seen unprecedented progress in improving generalization in machine vision driven by how algorithms process images through cutting-edge architectures and data augmentation techniques. Here, we present an alternative direction bringing together ideas from vision sciences, computer graphics, and computer vision---mimicking the human visual diet.

There is a long history of studies showing that changing the visual diet alters the visual cortex~\cite{kreiman2021biological, kandel2000principles}. Such developmental neuroscience studies include evaluating the consequences of rearing animals in visual environments deprived of binocular vision~\cite{hubel1964effects}, environmental directionality~\cite{daw1976kittens}, temporal contiguity~\cite{wood2018development}, surface features~\cite{wood2022development}, or faces~\cite{arcaro2017seeing}, among others. Some studies have also tried to quantitatively define the human visual diet~\cite{clerkin2017real}. In a similar vein, some computational works have also studied how the training data distribution (visual diet) impacts visual search assymetries~\cite{gupta2021visual}, and object recognition~\cite{madan2022and,madan2021small}. Here, we combined ideas from biological and computer vision and evaluated the impact of a human-like visual diet on the generalization behavior of visual recognition models. 

Through controlled analyses, we show that two hallmarks of human visual diet (transformational diversity and scene context) provide significant improvements in generalization across real-world transformations compared to existing popular approaches. These experiments are enabled by two computational contributions introduced here---a dataset which mimics the human visual diet, and an architecture which can leverage this visual diet. 

As a first step in mimicking the human visual diet, we focus on two important aspects of this diet---scene context and transformational diversity. There is a long history of work studying the role scene context in human vision~\cite{doi:10.1177/09567976211032676, greene2013statistics, graef1992scene, torralba2006contextual, henderson1992object, mathis1998does,zhang2020putting} and modeling contextual cues computationally~\cite{torralba2006contextual, taylor2013context,zhang2020putting,bomatter2021pigs}. Additinoally, recent work in machine learning has stressed the importance of data diversity for generalization~\cite{madan2021small, madan2022and}. These two aspects of the human visual diet are certainly not exhaustive and multiple other features warrant future investigation, including depth information, occlusions, and dynamic cues.  

The concept of generalization is ill-defined and depends on the training and test distributions. Humans also struggle with generalization when image statistics change (e.g., consider humans trying to read bar codes in the supermarket, learning to diagnose clinical images, or deciphering the shapes of new galaxies). A laudable goal in computer vision is often to align humans and machines. Such alignment takes the form of avoiding adversarial attacks that are imperceptible for humans but alter machine-produced labels, enhancing computer vision function in the real world, or even recognition challenges that would benefit from human-machine collaboration. Ensuring that the visual diets are similar can help accelerate such alignment. 

Advances in algorithms usually accompany progress in building better datasets. The introduction of a two-stream architecture constitutes a reasonable first-order approach to capitalize on transformational diversity and contextual cues. At the same time, we expect that more powerful datasets will incentivize the development of better algorithms that could also be in turn inspired and constrained by biological vision.


\section{Methods}\label{sec:methods}
\subsection{Datasets mimicking attributes of the human visual diet}\label{sec:datasets}
Several benchmarks have been proposed to study generalization in computer vision. On one hand, we have datasets like ImageNet-P~\cite{hendrycks2019benchmarking} and ImageNet-C~\cite{hendrycks2019benchmarking}, which introduce controlled synthetic noise that can be quantified, but the introduced noise does not reflect real-world transformations. On the other hand, there are domain generalization datasets like PACS~\cite{li2017pacs}, VLCS~\cite{torralba2011unbiased}, Office-Home~\cite{venkateswara2017officehome}, DomainNet~\cite{peng2019domainnet}, and Terra Incognita~\cite{beery2018recognition}, among others. While these datasets do include  transformations that exist in the real world, the distribution shift between domains in these benchmarks cannot be quantified or disentangled into scene parameters. For instance in PACS~\cite{li2017pacs}, it is unclear how the \textit{Photo} $\rightarrow$ \textit{Cartoon} domain shift differs from a \textit{Photo} $\rightarrow$ \textit{Art} shift. Furthermore, all these datasets  show objects with minimal context, as they are composed of crops centered around objects. The proposed HVD dataset was designed to address these issues---it shows objects in context, with controlled, disentangled real-world transformations. Below we introduce HVD, and also modify an existing natural image dataset to partly mimick the human-like visual diet.

~\subsubsection{Human Visual Diet (HVD) Dataset}\label{subsec:ssic}
We reconstructed 1,288 real-world scenes from the ScanNet dataset with the exact same 3D objects, scene layouts, class distributions and camera parameters using the OpenRooms framework~\cite{li2020openrooms, li2020inverse}. With these scenes, we created $15$ photo-realistic domains with $3$ types of real-world transformations including---lighting, material, and viewpoint changes. Some sample images are provided in\textbf{ Fig.~\ref{fig:fig_2_datasets}(a)}. We rendered $19,800$ images for each domain, which results in $70,000$ object instances from $13$ indoor object categories. Across all domains, this amounts to roughly $300,000$ images and $1$ million object instances. Below, we explain how different semantic shifts were created.

\textbf{Material shift domains:} We used $250$ high quality, procedural materials from Adobe Substances including different types of wood, fabrics, floor and wall tiles, and metals, among others. These were split into sets of 50 materials each to create $5$ different material domains (supplementary \textbf{Fig.~\ref{fig:material_shift}}). For each domain, its $50$ materials were randomly assigned to scene objects. One domain was held out for testing (OOD Materials), and never used for training any model.

\textbf{Light shift domains:} Outdoor lighting was controlled using $250$ High Dynamic Range (HDR) environment maps from the Laval Outdoor HDR Dataset~\cite{hold2019deep} and OpenRooms, which were split into $5$ sets of $50$ each (one set per domain). Disjoint sets of indoor lighting were created by splitting the HSV color space into chunks of disjoint hue values. Each domain sampled indoor light color and intensity from one chunk (supplementary \textbf{Fig.~\ref{fig:light_shift}}). One domain was held out for testing (OOD Light), and never used for training.

\textbf{Viewpoint shift domains:} Controlling object viewpoints presents a challenge as indoor objects are seen across a variety of azimuth angles (i.e., side vs front) across 3D scenes. Thus, to create disjoint viewpoint domains (supplementary \textbf{Fig.~\ref{fig:viewpoint_shift}}) we chose to control the zenith angle by changing the height at which the camera is focusing. Again, of the $5$ domains, one was held out for testing (OOD Viewpoints).



\subsubsection{Natural image test set from ScanNet}
To create the real-world test dataset, we sampled images from the ScanNet dataset~\cite{dai2017scannet}. ScanNet contains scenes captured using a moving camera from which frames can be extracted. We manually compared ScanNet scenes with our reconstructions of these scenes using OpenRooms, and selected 3D scenes for which OpenRooms was successfully able to recreate all three---scene layout, camera parameters, and geometry. As the video is continuous, nearby frames are highly similar. Therefore, we subsampled one frame for every 100 frames in the video. These frames were then annotated using LabelMe. To ensure a high quality test set, results are reported in the main paper on a subset of $350$ images which are not blurry and do not have significant clutter. However, here we report numbers on a larger subset of $700$ test images where we allow clutter and minor blurring so as to achieve a bigger test set. These results are reported in supplemtary \textbf{Sec.~\ref{supplesec:additional_controls}}. The same procedure was also followed to hand-annotate $8,000$ training object instances from the HVD dataset to ensure there is no spurious impact of the annotation procedure on the performance of models when tested on ScanNet.

~\subsubsection{Semantic-iLab dataset}\label{subsec:semantic_ilab}
To ensure our findings extend to natural images, we modified the iLab~\cite{borji2016ilab} dataset to create a natural image dataset with controlled variations in Lighting, Material, and Viewpoint, as shown in \textbf{Fig.~\ref{fig:fig_2_datasets} (b)}. We call this the Semantic-iLab dataset. The iLab dataset contains objects from $15$ categories placed on a turntable and photographed from varied viewpoints. A foreground detector was used to extract a mask for the object in each image. Material variations were implemented using AdaIN~\cite{huang2017arbitrary} based style transfer on these images and overlaying the style transferred image onto the object mask in the original image. Lighting changes were implemented by modifying the white balance. Note that unlike HVD, this dataset does not contain scene context. Additional details can be found in supplementary \textbf{Sec.~\ref{supplsec:semantic_ilab_images}}. 



\subsection{Human Diet Network (HDNet)}\label{sec:HDNet}
\begin{figure*}[!t]
    \centering
    \centering
    \subfloat[Context-aware Feature Extraction]{\includegraphics[width=0.45\linewidth]{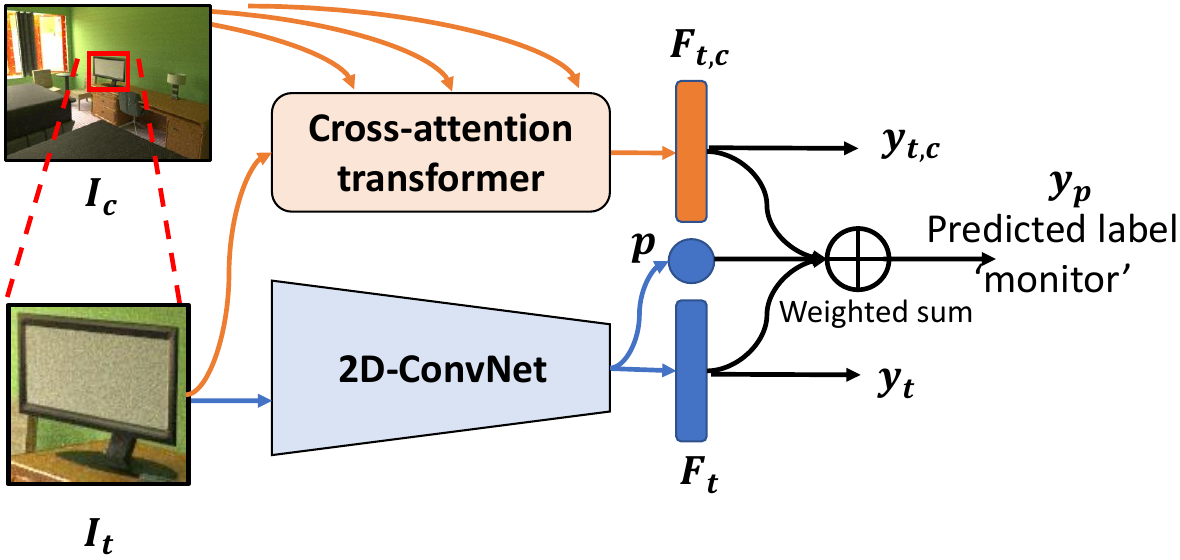}\label{fig:figs2a}}
    \subfloat[Contrastive Learning]{\includegraphics[width=0.3\linewidth]{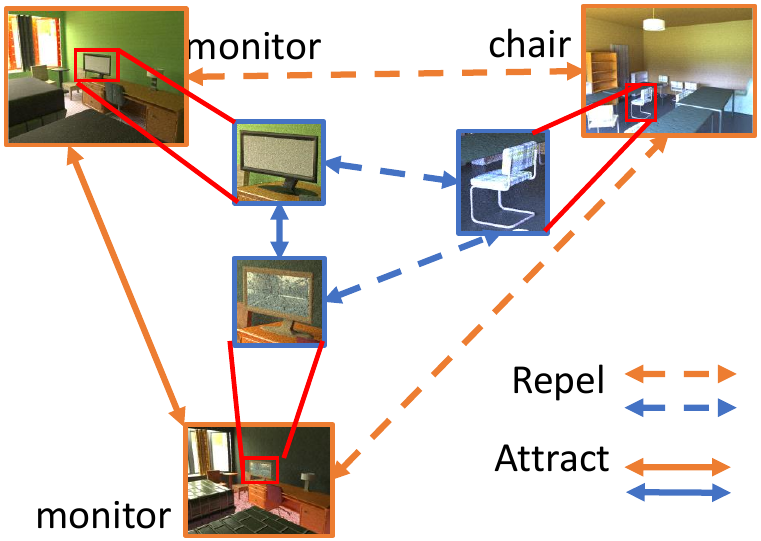}\label{fig:figs2b}}
    \caption{\textbf{Architecture overview for Human Diet Network(HDNet)}. (a) Modular steps carried out by HDNet in context-aware object recognition. HDNet consists of 3 modules: feature extraction, integration of context and target information, and confidence-modulated classification. HDNet takes the cropped target object $I_t$ and the entire context image $I_c$ as inputs and extracts their respective features. These feature maps are tokenized and information from the two streams is integrated over multiple cross-attention layers. HDNet also estimates a confidence score $p$ for recognition using the target object features alone, which is used to modulate the contributions of $F_t$ and $F_{t,c}$ in the final weighted prediction $y_p$. (b) To help HDNet learn generic representations across domains, we introduce contrastive learning on the context-modulated object representations $F_{t,c}$ in the embedding space. Target and context representations for objects of the same category are enforced to attract each other, while those from different categories are enforced to repel. Pairs for contrastive learning are generated using various material, lighting or viewpoint shifts (\textbf{Sec.~\ref{subsec:ssic}}).}
    \label{fig:fig_6_architecture}
\end{figure*}

A schematic of the proposed HDNet is shown in \textbf{Fig~\ref{fig:fig_6_architecture}}. We start with CRTNet as the backbone \cite{bomatter2021pigs} and introduce critical modification of contrastive learning described below to enable generalization across semantic shifts.\looseness=-1

\subsubsection{Feature Extraction in Context-aware Recognition using a Cross-attention Transformer}
The context-aware recognition model in \cite{bomatter2021pigs} achieved superior performance in in-context object recognition when the training and test data are from the same domain. Here, we used the same backbone and briefly introduce the network architecture below (see \cite{bomatter2021pigs} for implementation details).

Given the training dataset $D=\{x_i,y_i\}_{i=1}^n$, HDNet is presented with an image $x_i$ with multiple objects and the bounding box for a single target object location. $I_{i,t}$ is obtained by cropping the input image $x_i$ to the bounding box
whereas $I_{i,c}$ covers the entire contextual area of the image $x_i$. $y_i$ is the ground truth class label for $I_{i,t}$. In this subsection, we focus on extracting context and target features in the embedding space and omit the index $i$ for simplicity. Inspired by the eccentricity dependence of human vision, HDNet has one stream that processes only the target object ($I_t, 224 \times 224$), and a second stream devoted to the periphery ($I_c, 224 \times 224$) which processes the contextual area.  

The context stream is a transformer decoder, taking $I_c$ as the query input and $I_t$ as the key and value inputs. The network integrates object and context information via hierarchical reasoning through a stack of cross-attention layers in the transformer, extracts context-integrated feature maps $F_{t,c}$ and predicts class label probabilities $y_{t,c}$ within $C$ classes. 

A model that always relies on context can make mistakes under distribution shifts. Thus, to increase robustness, HDNet makes a second prediction $y_t$, using only the target object information alone. A 2D CNN is used to extract feature maps $F_t$ from $I_t$, and estimates the confidence $p$ of this prediction $y_t$. Finally, HDNet computes a confidence-weighted average of $y_t$ and $y_{t,c}$ to get the final prediction $y_p$. If the model makes a confident prediction with the object only, it overrules the context reasoning stage.\looseness=-1
\subsubsection{Supervised Contrastive Learning for Domain Generalization}
 
 Contrastive learning has benefited many applications in computer vision tasks (\textit{e.g.}, \cite{qian2021spatiotemporal,chen2020simple,snell2017prototypical,zhang2019variational,kim2021selfreg}). However, all these approaches require sampling positive and negative pairs from real-world data. To curate positive and negative pairs, image and video augmentations operate in 2D image planes or spatial-temporal domains in videos. 
 Here we introduce a contrastive learning method on 3D transformations. 
 
 Our contastive learning framework builds on top of the supervised contrastive learning loss \cite{khosla2020supervised}. Given the training dataset $D=\{x_i,y_i\}_{i=1}^n$, we randomly sample $N$ data and label pairs $\{x_k,y_k\}_{k=1}^N$. The corresponding batch pairs used for constrative learning consist of $2N$ pairs $\{\tilde{x}_l,\tilde{y}_l\}_{l=1}^{2N}$, where $\tilde{x}_{2k}$ and $\tilde{x}_{2k-1}$ are two views created with random semantic domain shifts of $x_k (k=1,...,N)$ and $\tilde{y}_{2k} = \tilde{y}_{2k-1} = \tilde{y}_k$. Domain shifts are randomly selected from a set of HVD domains specified during training. For example,  if $x_k$ is from a material domain, $\tilde{x}_{2k}$ and $\tilde{x}_{2k-1}$ could be images from the same 3D scene but with different materials. For brevity, we refer to a set of $N$ samples as a batch and the set of $2N$ domain-shifted samples as their multiviewed batch.\looseness=-1 
 
 Within a multiviewed batch, let $m \in M := \{1,...,2N\}$ be the index of an arbitrary domain shifted sample. Let $j(m)$ be the index of the other domain shifted samples originating from the same source samples belonging to the same object category, also known as the positive. Then $A(m) := M \backslash \{m\}$ refers to the rest of indices in $M$ except for $m$ itself. Hence, we can also define $P(m) := \{p \in A(m): \tilde{y}_p = \tilde{y}_m\}$ as the collection of indices of all positives in the multiviewed batch distinct from $m$. $|P(m)|$ is the cardinality. The supervised contrastive learning loss is:
 \vspace{-2mm}
\begin{multline}
  L_{contrast} = \sum_{m \in M} L_{m} = \sum_{m \in M} \frac{-1}{|P(m)|} \sum_{p \in P(m)} \log \frac{\exp(z_m \cdot z_p /\tau) }{\sum_{a\in A(m)} \exp (z_m \cdot z_a / \tau)}
\end{multline}

 Here, $z_m$ refers to the context-dependent object features $F_{m,t,c}$
 on $\tilde{x}_m$ after L2 normalization. The design motivation is to encourage HDNet to attract the objects and their associated context from the same category and repel the objects and irrelevant context from different categories. 
 
 As previous works have demonstrated the essential role of context in object recognition \cite{bomatter2021pigs,zhang2020putting}, contrastive learning on the context-modulated object representations enforces HDNet to learn generic category-specific semantic representations across various domains. $\tau$ is a scalar temperature value which we empirically set to $0.1$.    
 
 Overall, HDNet is jointly trained end-to-end with two types of loss functions: first, given any input $x_m$ consisting of image pairs $I_{m,c}$ and $I_{m,t}$, HDNet learns to classify the target object using the cross-entropy loss with the ground truth label $y_m$; and second, contrastive learning is performed with features $F_{m,t,c}$ 
 extracted from the 
 context streams:
  \begin{equation}\label{equ:jointloss}
L = \alpha L_{contrast, c,t} + L_{classi, t} + L_{classi, p} + L_{classi, c,t}
\end{equation}
Hyperparameter $\alpha$ is set to $0.5$ to balance the supervision from constrastive learning and the classification loss. Supplementary Table ~\ref{table:contrastive_ablation} shows that the contrastive loss introduced in HDNet results in improved performance across all real-world transformations. 

\subsection{Experimental Details}\label{sec:experiment_details}
\subsubsection{Baseline Architectures}\label{subsec:baselines}
HDNet was compared against several baselines presented below. All models were trained on NVIDIA Tesla V100 16G GPUs. Optimal hyper-parameters for benchmarks were identified using random search, and all hyper-parameters are available in the supplement in \textbf{Sec.~\ref{supplesec:hparams}}.

\textbf{2D feed-forward object recognition networks:} Previous works have tested popular object recognition models in generalization tests \cite{geirhos2018generalisation,boyd2022human}. We include the same popular architectures ranging from 2D-ConvNets to transformers:
DenseNet \cite{iandola2014densenet}, ResNet \cite{he2016deep}, and ViT \cite{dosovitskiy2020image}. These models do not use context, and take the target object patch $I_t$ as input.

\textbf{Domain generalization methods:} We also compare HDNet to an array of state-of-the-art domain generalization methods (\textbf{Table~\ref{table:dg_benchmarks}}). These methods also use only the target object, and do not use contextual information. 

\textbf{Context-aware recognition models:} To compare against models which use scene context, we include CRTNet \cite{bomatter2021pigs} and Faster R-CNN \cite{ren2015faster}. CRTNet fuses object and contextual information with a cross-attention transformer to reason about the class label of the target object. We also compare HDNet with a Faster R-CNN \cite{ren2015faster} model modified to perform recognition by replacing the region proposal network with the ground truth location of the target object.  

\subsubsection{Evaluation of computational models}

Performance for all models is evaluated as the Top-1 classification accuracy. Error bars reported on all figures refer to the variance of per-class accuracies of different models. For statistical testing, p-values were calculated using a two-sample paired t-test on the per-category accuracies for different models. The t-test checks for the null hypothesis that these two independent samples have identical average (expected) values. For ScanNet, a t-test is not optimal due to the smaller number of samples, and thus a Wilcoxon rank-sum test was employed for hypothesis testing as suggested in past works~\cite{de2019using, posten1982two}. All statistical testing was conducting using the python package \textit{scipy}, and the threshold for statistical significance was set at $0.05$.

\section*{Data and Code Availability Statement} 
Source code and data are available at \url{https://github.com/Spandan-Madan/human_visual_diet}.

\section*{Acknowledgements}
This research is supported by the National Research Foundation, Singapore under its AI Singapore Programme (AISG Award No: AISG2-RP-2021-025), its NRFF award NRF-NRFF15-2023-0001, Mengmi Zhang's Startup Grant from Agency for Science, Technology, and Research (A*STAR), Mengmi Zhang's Startup Grant from Nanyang Technological University, and Early Career Investigatorship from Center for Frontier AI Research (CFAR), A*STAR. This work has also been partially supported by NSF grant IIS-1901030.

\bibliographystyle{unsrt}
\bibliography{literature}

\clearpage

\appendix
\section*{Appendix}

\appendix
\renewcommand{\thefigure}{Sup\arabic{figure}}
\renewcommand{\thetable}{Sup\arabic{table}}
\renewcommand{\theequation}{Sup\arabic{equation}}

\setcounter{figure}{0}
\setcounter{table}{0}
\setcounter{equation}{0}

\section{Sample images from the HVD Dataset}\label{supplsec:ssic_images}
\begin{figure*}[p]
    \centering
    \includegraphics[scale=0.25]{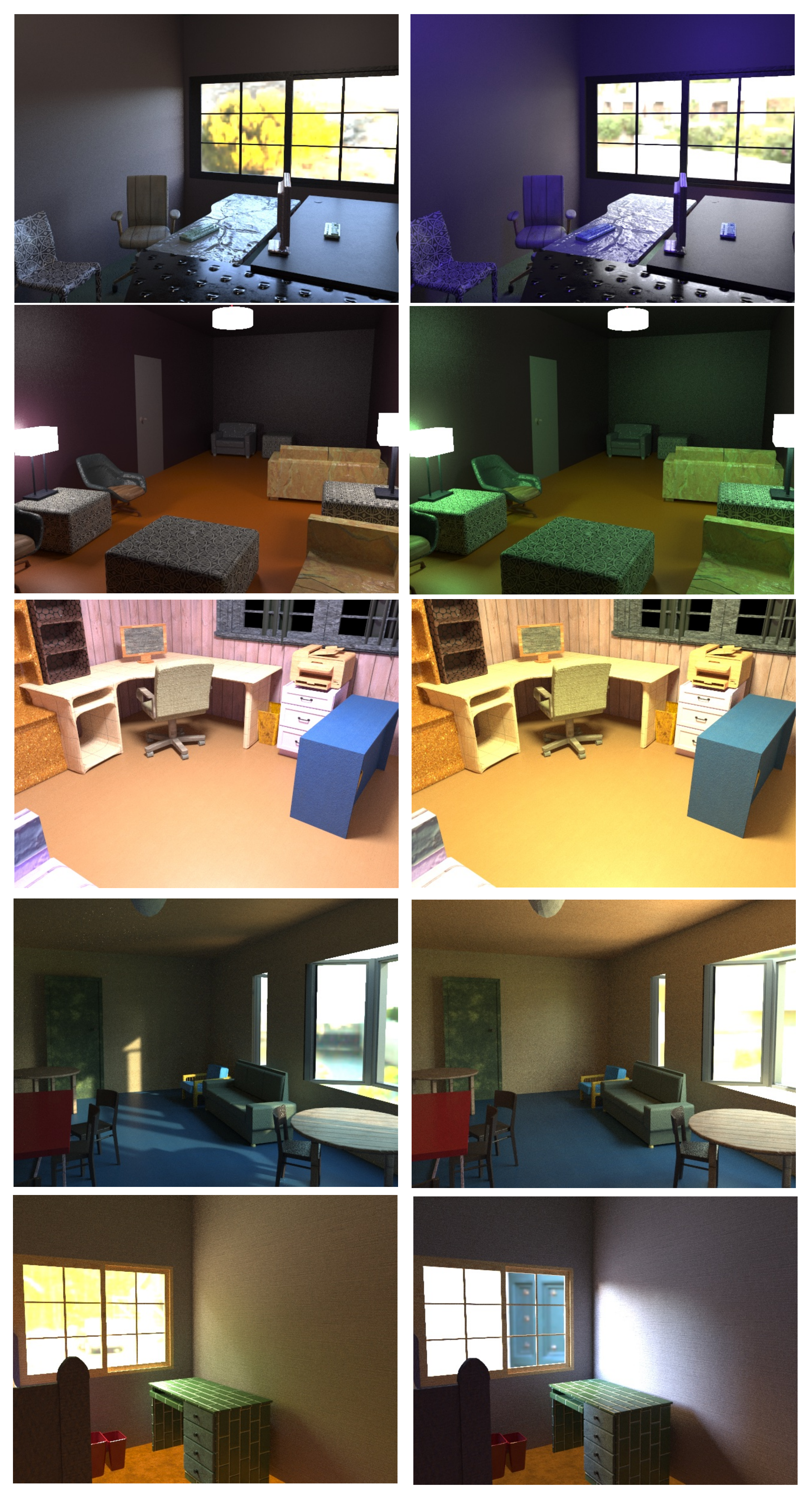}
    \caption{\emph{Example images showing lighting tranformations.} We show paired images from different lighting transformation domains between the right and left column in each row. All other parameters held constant.}
    \label{fig:light_shift}
\end{figure*}

\begin{figure*}[p]
    \centering
    \includegraphics[scale=0.25]{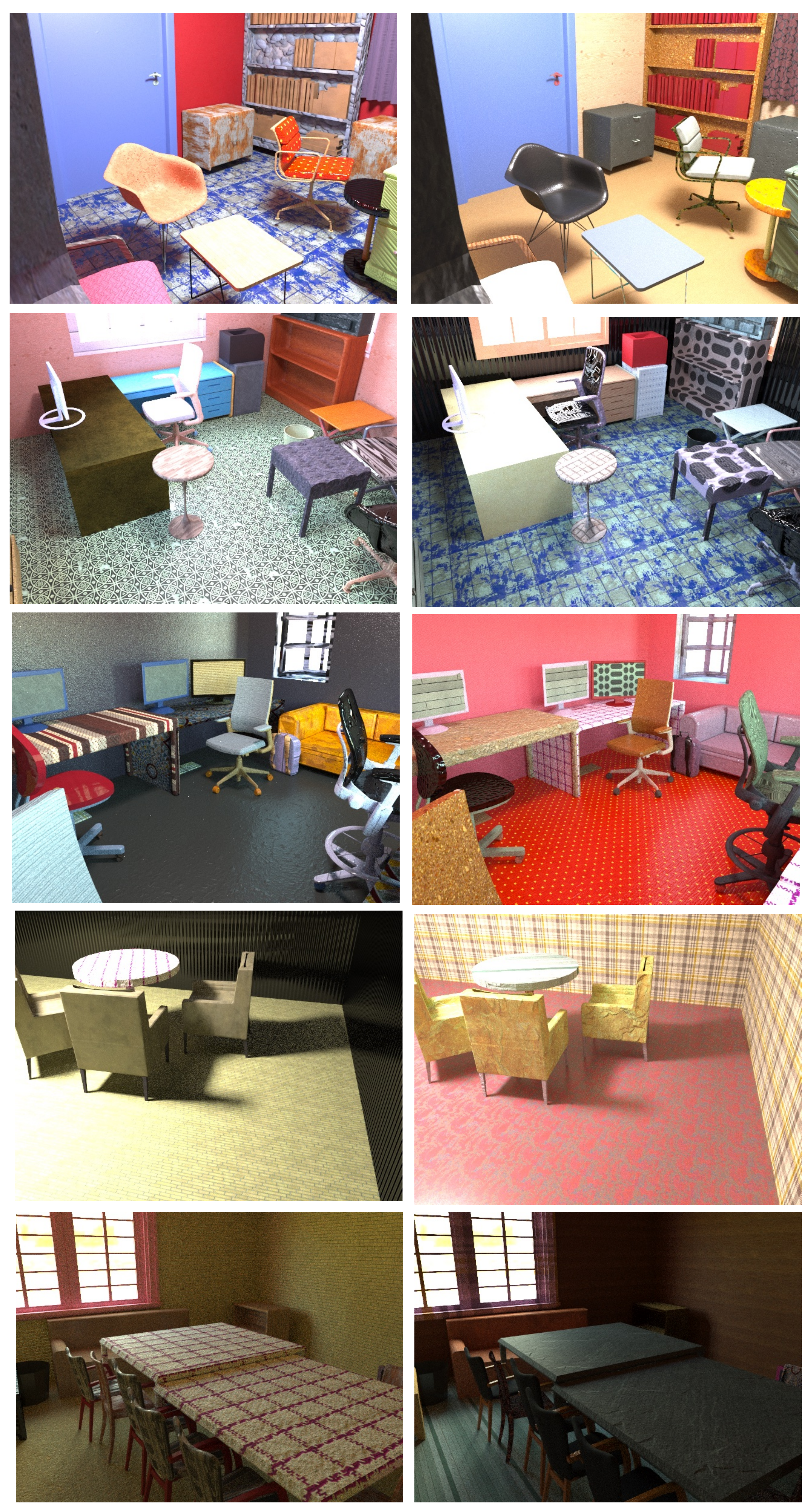}
    \caption{\emph{Example images showing material tranformations.} We show paired images from different material transformation domains between the right and left column in each row. All other parameters held constant}
    \label{fig:material_shift}
\end{figure*}

\begin{figure*}[p]
    \centering
    \includegraphics[scale=0.25]{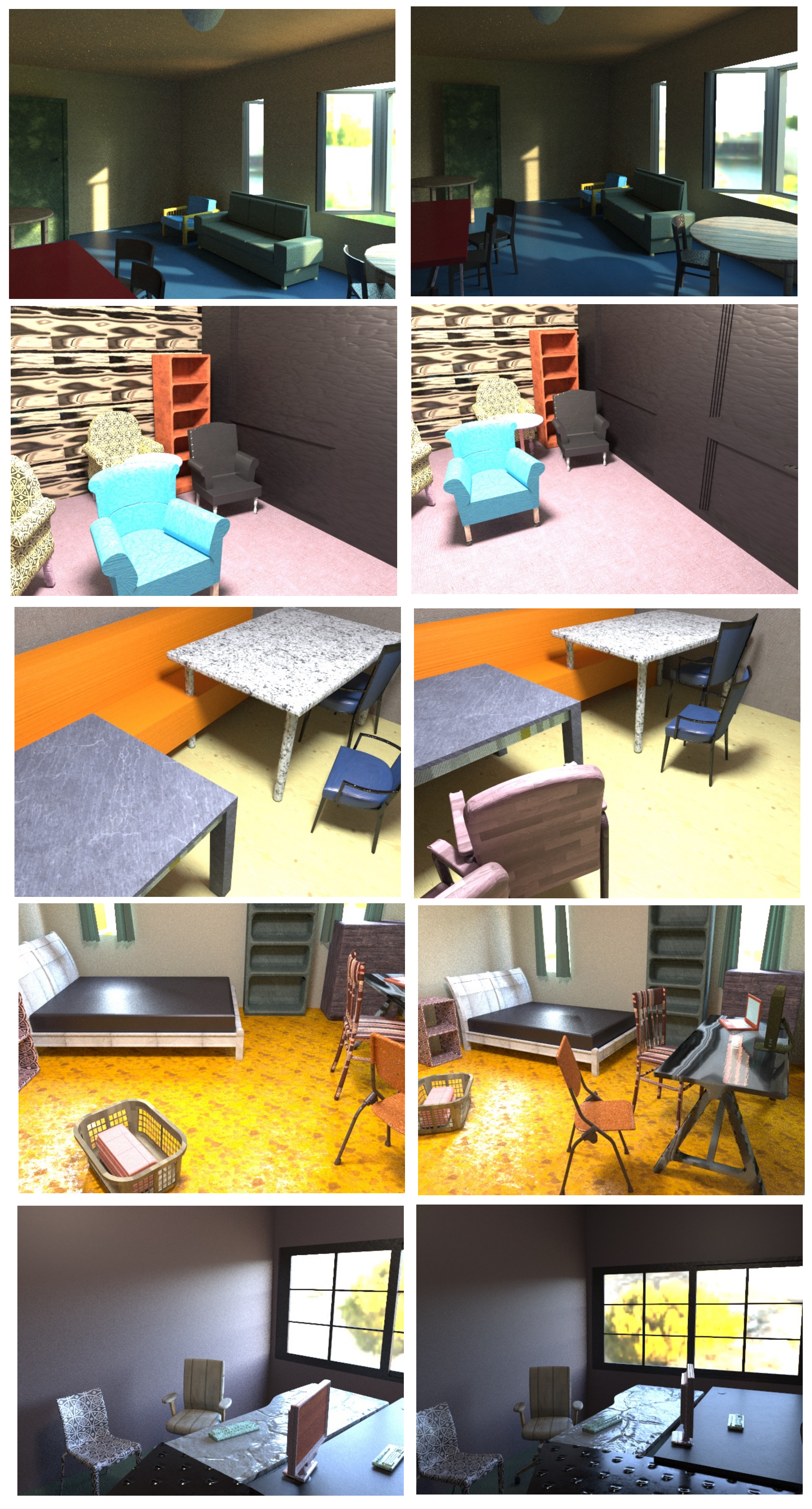}
    \caption{\emph{Example images showing viewpoint tranformations.} We show paired images from different viewpoint transformation domains between the right and left column in each row. All other parameters held constant}
    \label{fig:viewpoint_shift}
\end{figure*}

We present additional images from the HVD dataset. Each figure shows change in one scene parameter, while holding all others constant. In Fig.~\ref{fig:light_shift} we show images from two different light domains. Note that the first three rows in Fig~\ref{fig:light_shift} show different indoor lighting conditions controlled using indoor light color and intensity sampled from disjoint chunks of the HSV space. The last two rows show different outdoor lighting settings created by changing the environment maps. Similarly, Fig.~\ref{fig:material_shift} shows five different scenes from two training domains with a material shift. Fig.~\ref{fig:viewpoint_shift} shows viewpoint shifted domains. 

\section{Additional details for the creation of the Semantic iLab dataset}\label{supplsec:semantic_ilab_images}
\begin{figure*}
     \centering
    \includegraphics[width=\linewidth]{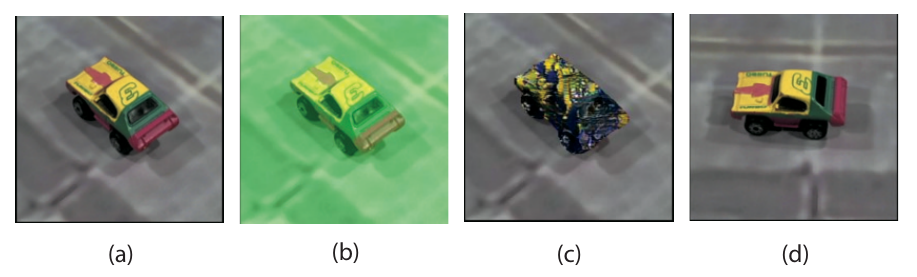}
    \caption{\textbf{Sample images from the Semantic-iLab dataset.} (a) Original image of a toy car placed on a turntable from the original iLab~\cite{borji2016ilab} dataset. (b) Light shifted image from \textit{Semantic-iLab} created by modifying the white balance of the original image. (c) Material shifted image from \textit{Semantic-iLab} created by modifying the original image by first detecting the foreground object mask, and overlaying the style transferred image on this mask. (d) Viewpoint shifted image of the same object from the iLab dataset.}
    \label{fig:semantic_ilab}
\end{figure*}

We show sample images from the \textit{Semantic iLab} dataset in Fig.~\ref{fig:semantic_ilab} created by modifying the existing iLab~\cite{borji2016ilab} dataset. This is a multi-view dataset, and hence already contains viewpoint shifted variations of the same objects. We modify the dataset to also contain material and light shifts. To mimick light shift, we modified the white balance of the original images, as shown in Fig.~\ref{fig:semantic_ilab}(b). For material shifts, we first run a foreground detector on these objects using Google's Cloud Vision API. We also run style transfer on these images using AdaIn~\cite{huang2017arbitrary}. Then, we overlay the style transferred image on to the object mask on the original image to mimick material shifts. Note that this is approximate, and does not model the physics of material transfer in the same way as our rendered HVD dataset which is far more photorealistic, as shown in Fig.~\ref{fig:material_shift}. Material shifted \textit{Semantic iLab} images are shown in Fig.~\ref{fig:semantic_ilab}(c). As the dataset is originally multi-view, we do not need to generate new viewpoints and can use images of a different viewpoint from the original dataset as shown in Fig.~\ref{fig:semantic_ilab}(d).

\section{HDNet ablations with contrastive loss}\label{supplesec:additional_controls}
We evaluate the contribution of the contrastive loss by training variations of HDNet on HVD with and without the contrastive loss as shown in Eq.~\ref{equ:jointloss}. These numbers are reported in Table~\ref{table:contrastive_ablation}. As can be seen, adding a contrastive loss improves performance for all three semantic shifts, providing evidence for its utility.

\begin{table}[h]
\centering
\begin{tabular}{c|c|c}
\hline
\hline
\multirow{3}{*}{\parbox{1.5cm}{\centering Semantic Shift}} & \multirow{3}{1.7cm}{\centering Without Contrastive Loss} & \multirow{3}{1.7cm}{\centering With Contrastive Loss} \\
& &\\
& &\\
\hline
 Viewpoint & 0.79 & \textbf{0.82}\\
 \hline
 Material & 0.89 & \textbf{0.94}\\
 \hline
 Lighting & 0.98 & \textbf{0.98}\\
 \hline
\end{tabular}
\caption{\textbf{Impact of removing contrastive loss}. We evaluate the contribution of the contrastive loss by training and testing HDNet on  the HVD dataset with and without the contrastive loss. The contrastive loss results in an improvement across all three semantic shifts.\label{table:contrastive_ablation}}
\end{table}

\section{Additional experiment for the role of context}
\begin{table}[h]
\centering
\begin{tabular}{c|c|c}
\hline
\hline
\multirow{3}{*}{\parbox{1.5cm}{\centering Semantic Shift}} & \multirow{3}{1.2cm}{\centering Target only} & \multirow{3}{1.7cm}{\centering Target and Context} \\
& &\\
& &\\
\hline
 Viewpoint & 0.77 & \textbf{0.82}\\
 \hline
 Material & 0.85 & \textbf{0.94}\\
 \hline
 Lighting & 0.97 & \textbf{0.98}\\
 \hline
\end{tabular}
\caption{\textbf{Training a two-stream HDNet with only target information}. As a third control for confirming the role of context, we train HDNet where both streams are passed just the target object. Thus, it is forced to learn without scene context. This results in a drop in performance for all semantic shifts, providing further evidence in support of the utility of scene context.\label{table:control_3}}
\end{table}
Besides results on the role of context presented in  \textbf{Table ~\ref{table:control_1}} and\textbf{ Table ~\ref{table:control_2}}, we present here an additional control evaluating the contribution of scene context on generalization. For this, we train HDNet such that both streams are trained with the target object. Thus, this modified version is forced to learn without scene context. These results are shown in Table.~\ref{table:control_3}. For all semantic shifts, forcing HDNet to learn with only the target results in a drop in accuracy. This provides further evidence supporting the utility of scene context in enabling generalization. 

\subsection{Results with  a larger, less controlled ScanNet test set.}
\begin{table*}
\centering
\begin{tabular}{c|c|c|c|c|c|c|c|c|c|c||c}
    \hline
    \hline
    \multirow{3}{*}{\parbox{1cm}{\centering Test Dataset}} & 
    \multirow{3}{*}{\parbox{1cm}{ResNet ~\cite{he2016deep}}} & 
     \multirow{3}{*}{\parbox{0.5cm}{ViT ~\cite{dosovitskiy2020image}}} & \multirow{3}{*}{\parbox{0.6cm}{AND Mask \cite{shahtalebi2021sand}}}& \multirow{3}{*}{\parbox{0.6cm}{CAD \cite{blanchard2017domain}}} & \multirow{3}{*}{\parbox{0.6cm}{COR AL \cite{DBLP:journals/corr/SunS16acoral}}} & \multirow{3}{*}{\parbox{0.6cm}{ERM \cite{vedantam2021empirical}}} & \multirow{3}{*}{\parbox{0.6cm}{IRM \cite{arjovsky2019invariant}}} & \multirow{3}{*}{\parbox{0.6cm}{MTL \cite{blanchard2017domainmtl}}}  &  \multirow{3}{*}{\parbox{0.6cm}{Self Reg \cite{kim2021selfreg}}} & \multirow{3}{*}{\parbox{0.8cm}{VREx \cite{krueger2021out}}} & \multirow{3}{*}{\parbox{1cm}{\centering{HDNet (ours)}}} \\
     & & & & & & & & & & &\\
     & & & & & & & & & & &\\
     \hline
      ScanNet & 0.35 & 0.29 & 0.43 & 0.40 & 0.42 & 0.48 & 0.46 & 0.46 & 0.53 & 0.42 & \textbf{\textbf{0.61}}\\
      \hline
\end{tabular}
\caption{\textbf{Human visual diet improves generalization to larger real world dataset as well}. We curated a larger subset of ScanNet images, allowing more complex real world scenarios like blurry images, clutter and occlusions. We report the capability of models to generalize from synthetic HVD images to this more complex subset of ScanNet. HDNet leveraging human-like visual-diet outperforms all baselines on this more complex dataset as well.}\label{table:real_world_generalization_larger_dataset}
\end{table*}

We extend the generalization to real-world results presented in the main paper by reporting these numbers on a larger test set created by annotating additional images from ScanNet. As ScanNet was created by shooting video footage of 3D scenes, many frames can be blurry. In the original, smaller test-set such blurry frames were removed to ensure a higher quality test set. However, here we also include additional images with lower fidelity to report numbers on a larger test set. These numbers are reported in Table.~\ref{table:real_world_generalization_larger_dataset}. The trend is consistent with results reported on a smaller, more controlled subset in the main paper---HDNet outperforms all other benchmarks by a large margin. As expected, including these images in the test set results in a drop in accuracy across all methods. All models were trained on synthetic images from HVD and were tested on a test set of natural images from ScanNet.

\section{Hyperparameters}\label{supplesec:hparams}

\textbf{HDNet:} As our model builds on top of CRTNet~\cite{bomatter2021pigs} as backbone, we use the same hyperparameters for the backbone as reported in the original paper. All models were trained for 20 epochs with a learning rate of 0.0001, with a batch size of 15 on a Tesla V100 16Gb GPU.

\textbf{Domain generalization: }We used the code from Gulrajani et al.~\cite{gulrajani2020search} to train and test domain generalization methods on our dataset. The code is available here: \url{https://github.com/facebookresearch/DomainBed}. To begin, we ran all available models and tried 10 random hyperparameter initializations. Of these, we picked the best performing hyperparameter seed---24596. We also picked the top performing algorithms as the baselines reported in the paper.

\textbf{FasterRCNN:} We used the code from Bomatter et al.~\cite{bomatter2021pigs} to train and test the modified FasterRCNN model for recognition. The code is available here: \url{https://github.com/kreimanlab/WhenPigsFlyContext}, and we used the exact hyperparameters mentioned in the repo.

\end{document}


\maketitle
In this document we provide additional details and experiments which were not presented in the main submission for the sake of brevity. This document follows the same order as the main paper.

\fi